\documentclass[11pt,a4paper]{article}
\usepackage[hyperref]{emnlp2020}
\usepackage{times}
\usepackage{latexsym}
\usepackage{algorithmicx}
\usepackage{algpseudocode}
\usepackage{algorithm}
\usepackage{multirow}
\usepackage{amsmath,graphicx}

\usepackage{enumitem}
\usepackage{microtype}

\aclfinalcopy 

\newcommand{\OURMODEL}{\textcolor{black}{\textsc{Megatron-Cntrl}}}
\newcommand{\OurIIVIIm}{\textcolor{black}{\textsc{Megatron-Cntrl-124m}}}
\newcommand{\OurIIIIVVm}{\textcolor{black}{\textsc{Megatron-Cntrl-355m}}}
\newcommand{\OurVIIVIVIIIm}{\textcolor{black}{\textsc{Megatron-Cntrl-774m}}}
\newcommand{\OurIIb}{\textcolor{black}{\textsc{Megatron-Cntrl-2b}}}
\newcommand{\OurVIIIb}{\textcolor{black}{\textsc{Megatron-Cntrl-8b}}}

\newcommand{\MODEL}{\textcolor{black}{\textsc{M-Cntrl}}}
\newcommand{\IIVIIm}{\textcolor{black}{\textsc{M-Cntrl-124m}}}
\newcommand{\IIIIVVm}{\textcolor{black}{\textsc{M-Cntrl-355m}}}
\newcommand{\VIIVIVIIIm}{\textcolor{black}{\textsc{M-Cntrl-774m}}}
\newcommand{\IIb}{\textcolor{black}{\textsc{M-Cntrl-2b}}}
\newcommand{\VIIIb}{\textcolor{black}{\textsc{M-Cntrl-8b}}}
\newcommand{\ANTONYM}{\textcolor{black}{\textsc{Ant}}}

\newcommand{\OurANTONYM}{\textcolor{black}{\textsc{Megatron-Cntrl-Ant}}}

\newcommand{\KEYWORD}{k} 
\newcommand{\KNOWTRIP}{t} 
\newcommand{\UniSenEnc}{U}
\newcommand{\UniSenEncFull}{\textsc{Use}}
\newcommand{\RAKE}{\textsc{Rake}}
\newcommand{\RetKnowTrip}{R} 
\newcommand{\GPTTWO}{\textsc{Gpt-2}}

\newcommand{\KNOWBASE}{G} 
\newcommand{\BERT}{\textsc{ Bert}}

\newcommand{\ROC}{\textsc{ Roc}}

\title{\textcolor{black}{\OURMODEL:} Controllable Story Generation with External Knowledge Using Large-Scale Language Models}

\author{Peng Xu$^{\ddagger}$\thanks{\hspace{1mm} This work was done during the internship of Peng Xu at NVIDIA. Corresponding authors: Peng Xu, Mostofa Patwary.}, Mostofa Patwary$^{\mathsection}$, Mohammad Shoeybi$^{\mathsection}$, Raul Puri$^{\mathsection}$, \\
  \textbf{Pascale Fung$^{\ddagger}$, Anima Anandkumar$^{\mathsection}$, Bryan Catanzaro$^{\mathsection}$} \\
  $^{\ddagger}$The Hong Kong University of Science and Technology,
  $^{\mathsection}$NVIDIA \\
  \texttt{peng.xu@connect.ust.hk, mpatwary@nvidia.com
  } 
  
  }

\date{}

\begin{document}
\maketitle
\begin{abstract}
Existing pre-trained large language models have shown unparalleled generative capabilities. However, they are not controllable. In this paper, we propose \textcolor{black}{\OURMODEL}, a novel framework that uses large-scale language models and adds control to text generation by incorporating an external knowledge base. Our framework consists of a keyword predictor, a knowledge retriever, a contextual knowledge ranker, and a conditional text generator. As we do not have access to ground-truth supervision for the knowledge ranker, we make use of weak supervision from sentence embedding. The empirical results show that our model generates more fluent, consistent, and coherent stories with less repetition and higher diversity compared to prior work on the ROC story dataset.
We showcase the controllability of our model by replacing the keywords used to generate stories and re-running the generation process. Human evaluation results show that 77.5\% of these stories are successfully controlled by the new keywords. Furthermore, by scaling our model from 124 million to 8.3 billion parameters we demonstrate that larger models improve both the quality of generation (from 74.5\% to 93.0\% for consistency) and controllability (from 77.5\% to 91.5\%). 


%
\end{abstract}

\section{Introduction}


\begin{table}[t]
\centering\scalebox{0.71}{
\begin{tabular}{|p{2.9cm}@{\hskip3pt}|@{\hskip3pt}p{7.cm}|}
\hline
  Context
  &  [FEMALE] was on a long road trip .   \\ \hline
 {\small \citet{guan2020knowledge}}
  & she had just driven into town . she had been driving for hours . she was so tired she decided to take a nap . when she woke up , she still had n't slept well . \\ \hline  
 {\OurIIVIIm} {\color{blue}{keyword: driving}} & {she was on her way to see a strange country.  she had never seen anything like that before.  finally she saw a small town that was so small.  it was a great little town with a nice park !} \\ \hline
 {\OurVIIIb} {\ \ \ \ \ \ \ \ \ \ \ \ \ } {\color{blue}keyword: driving} & she was driving in the middle of [PLACE]. all of a sudden the tire pressure light came on. she got out to check her tire. it was flat so she used the roadside assistance. \\ \hline
 {\OurVIIIb}-{\ANTONYM} {\color{blue}keyword: attract} &  she really wanted to see a few attractions.  the first one she saw was a giant water park.  it was amazing.  it ended up being a fun experience. \\ \hline  
\end{tabular}}
\caption{\label{tab:example_0} Stories  generated by models of increasing capacity and controllability. As the model size grows, story quality becomes increasingly coherent, fluent, and logically consistent. The last row demonstrates how {\OurVIIIb-{\ANTONYM}} model controls the story generation with a new keyword, ``attract". Note that {[MALE] and [FEMALE] denote names and [PLACE] denotes locations.} }
\vspace{-4mm}
\end{table}


Text generation has recently attracted significant attention from the research community as large pre-trained language models, such as {\GPTTWO} \cite{radford2018improving, radford2019language} demonstrated promising results for generating long, grammatically correct, and fluent text. Finetuning these models has shown significant improvements in downstream tasks, such as  persona chat~\cite{wolf2019transfertransfo}. However, one non-negligible drawback of these large models is the lack of knowledge which humans use to produce natural text. For example, {\GPTTWO} based models produce degraded generations that are illogical and ungrammatical for knowledge-driven generation tasks, such as story generation. \citet{guan2020knowledge} therefore introduced commonsense knowledge to the pre-trained language model by further finetuning on commonsense datasets. Although implicit encoding of knowledge is helpful for knowledge incorporation, there is still a lack of training mechanism to teach the model when and what to incorporate from external knowledge.

\begin{figure*}[t]
\begin{center}
    \scalebox{0.8}{
        \includegraphics[width=\linewidth]{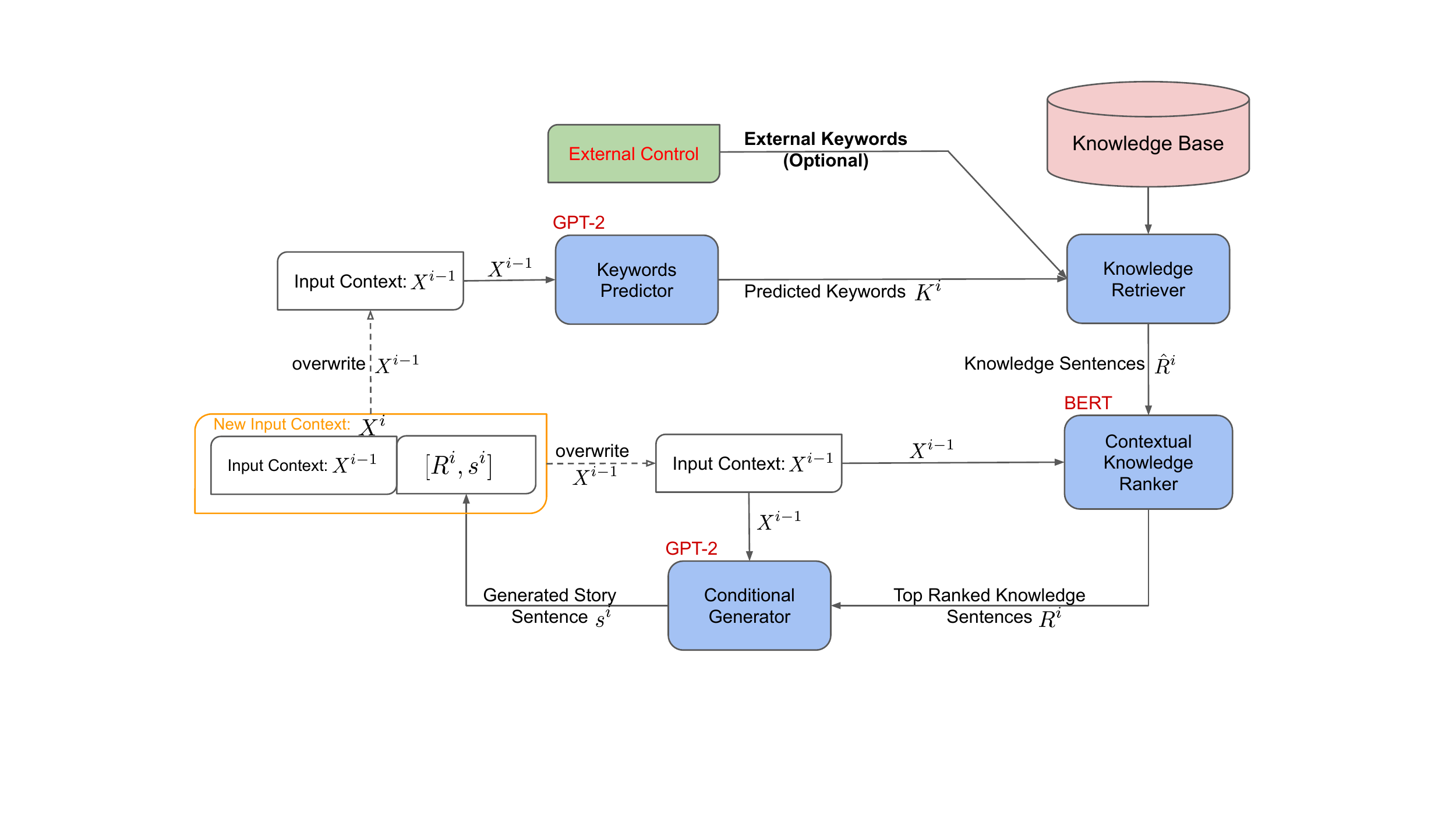}
    }
\end{center}
\vspace{-2mm}
\caption{Overview of our generation process. Based on an input context, we generate keywords for future context, use the keywords to retrieve the relevant knowledge from an external knowledge-base, filter them based on their relevance to the context, and use the top scored knowledge sentences to guide the generation.}
\label{fig:sys_diagram}  
\vspace{-3mm}
\end{figure*}

In addition, these large pre-trained language models are hard to control. Recently, plug-and-play language models ~\citet{dathathri2019plug} addressed whole document controllability by adding a linear classifier on top of {\GPTTWO} to predict whether generated text observes a particular style or property. 
\citet{keskar2019ctrl} controlled a 1.2B parameter language model generation via the use of control codes prepended to the model input. 
\citet{boyd2020large} controlled the personality of a dialogue agent by conditioning it on prior conversations of a target actor. However, these controlling conditions are predefined, limited in their capability, and are only used once at the beginning to condition the generation of the rest of the document. They do not provide control granularity at either a sentence or sub-document level.

In this work, we address these shortcomings and develop an efficient controllable text generation framework that we apply to the story generation task. In order to provide manual control to users through a set of interpretable keywords, we first develop a keyword \textcolor{black}{predictor} model for the next sentence.
These keywords are then used to retrieve knowledge sentences from an external knowledge base. 
Not all the retrieved knowledge is relevant to the story context and often it is noisy. To this end, we introduce a novel contextual ranker that ranks knowledge sentences based on the relevance to the context.  As we do not have access to ground-truth supervision for this contextual knowledge ranker, we make use of sentence embedding for weak supervision. The top-ranked knowledge sentences from the knowledge ranker are then fed to the conditional text generator to guide generation. By giving the knowledge in addition to the context, we provide rich information for the generator to attend to and help the model better understand the rationale between sentences. 
Table \ref{tab:example_0} shows an example of controllable story generation with increasing model capacity.

\paragraph{\bf Summary of Contributions:} 
\begin{itemize}[leftmargin=*]
    \item We propose a novel generation framework that allows dynamical incorporation of external knowledge into language model as well as control for text generation. 
    \item Using both automatic metrics as well as human evaluations, we demonstrate that our model generates more fluent, consistent, and coherent stories with lower repetition rate and higher diversities compared to the previous state-of-the-art on {\ROC} story datasets \cite{mostafazadeh2016story}.
    \item We showcase the controllability of our model by replacing  the  keywords  used  to  generate  stories. Human evaluation results show that up to 91.5\% of the generated stories are successfully controlled by the new  keywords .
    \item We scale our model from 124 million to 8.3 billion parameters and demonstrate that both qualities, as well as controllability of the generations, improve as the model size increases.
    
\end{itemize}

\section{Framework}
\label{sec:framework}

In our problem setup, we complete a story using the first sentence as input, similar to \citet{guan2020knowledge}. We augment the generation process with an external knowledge-base and develop a methodology that can guide and control the story generation. Our approach consists of the following four steps connected together as shown in Figure \ref{fig:sys_diagram}:

\begin{enumerate}
    \item Given the story context, a keyword \textcolor{black}{predictor} model first predicts a set of keywords for the next sentence yet to be generated.
    \item A knowledge retriever then takes the generated keywords and queries an external knowledge-base where each knowledge triple is converted into natural language ``knowledge sentences" using templates. 
    \item A contextual knowledge ranker then ranks the external knowledge sentences based on their relevance to the story context.
    \item Finally, a generator takes both the story context as well as the top-ranked knowledge sentences as input and generates the next sentence in the story. The output sentence is appended to the story context and steps 1-4 are repeated.
\end{enumerate}
This formulation naturally allows controllability by replacing the keyword \textcolor{black}{prediction} process with manual external keywords. \textcolor{black}{This work uses dynamic planning of the keywords and knowledge at each generation step. This allows the users to participate and control the generation on the go. As a result, they don’t need to pre-specify the keywords explicitly. We also note that it is challenging to statically plan all the knowledge needed for generation at the beginning. This issue becomes severe for long generations.} To formalize this method, we start by introducing notation used throughout the paper and then detail each aforementioned four steps in the following subsections.

{\bf Notation:} A knowledge-base, ${\KNOWBASE}$ is defined as a set of knowledge triples $t=(\text{subject}, \text{relation}, \text{object})$.  A knowledge sentence, $r$ is defined as $r=T(t)$ by mapping $t$ using predefined templates $T$. For example, \textit{(eiffel tower, AtLocation, paris)} is transformed into {\it eiffel tower is at paris}. \textcolor{black}{We should highlight that since our framework transforms the triple knowledge database into natural language sentences, any knowledge base in natural language format can be readily incorporated into our framework.} We use superscripts to index \textit{story sentences} and define a story $S$ of length $l$ as a sequence of individual story sentences $s^i$ where $S={\{s^1, s^2, \cdots, s^l}\}$. We use $K^i=\{k_1^i, \cdots, k_q^i\}$ to denote the keywords associated with story sentence $s^i$. A keyword $k_q^i$ is made up of subword tokens from our language model's vocabulary. Note that the number of keywords $q$ per sentence varies and can be zero. We define $R^i=\{r_1^i, \cdots, r_v^i\}$ as the knowledge associated with $s^i$, where $r_j^i$ denotes the $j$-th \textit{knowledge sentence} associated  $s^i$. The number of knowledge sentences $v$ varies per sentence and can be zero. Note that $v\neq q$ because a keyword can have multiple knowledge triples associated with it. Given this notation, we define the story context $X^i=\{x^1, \cdots, x^i\}$ where $x^i=[R^i, s^i]$.  The goal of this work is to generate $x^i$ given $X^{i-1}$, that is to first predict the knowledge $R^i$ contained in $s^i$  and then predict $s^i$ itself.

\subsection{Keyword \textcolor{black}{Predictor} Model}
\label{sec:key-gen}
To provide manual control to users, we first develop a keyword \textcolor{black}{predictor} model. Given the current story context $X^{i-1}$, the model predicts a set of keywords $K^i$ for the next sentence yet to be generated. The prediction of keywords instead of directly predicting knowledge triples not only allows us to control the generation in an interpretable manner, but it also helps to greatly reduce the search space for the knowledge triples. We formulate this keyword \textcolor{black}{prediction} problem similar to a left-to-right language model where the goal is to predict the string of concatenated keywords:
\begin{equation}
    p(K^i | X^{i-1}) = \prod_{j=1}^q p(k_j^i | X^{i-1}, K^i_{<j}),
\end{equation}
where $K_{<j}$ denotes all the predicted keywords up to the $j$th keyword and $p$ is the probability distribution. We use a {\GPTTWO} \cite{radford2019language} transformer to model this probability distribution.
We optimize the keyword \textcolor{black}{predictor} with maximum likelihood training and a next token prediction loss. Following \citet{yao2019plan}, we provide the labels for $K^i$ by extracting keywords from a ground truth training sentence $s^i$ using the {\RAKE} algorithm \cite{rose2010automatic} to train our keyword \textcolor{black}{predictor. Note that our model allows generation of multiple keywords and thus provides the flexibility to choose a subset of them as the control signal to fit in the generation.}

\subsection{Knowledge Retrieval}
\label{sec:know-ret}

In this step, we use the generated keywords $K^i$ in Section \ref{sec:key-gen} and retrieve all the related knowledge triples from our knowledge base $G$. This is simply done by converting all knowledge triples into knowledge sentences using predefined templates and then matching keywords against the knowledge sentences. This results in the knowledge set $\hat{R}^i=\{\hat{r}_1^i, \cdots, \hat{r}_z^i\}$ with size $z$. Future work will focus on replacing this simple retrieval with a learnable module similar to \citet{REALM}.

\begin{algorithm}[h]
{\fontsize{10.1pt}{10.1pt}\selectfont
\caption{Building Pseudo Label of $R^i$}
\label{alg:kb_retrieval}
\textbf{Input:} Story sentence $s^i$ and its preceding sentence $s^{i-1}$, {\UniSenEncFull} encoder $\UniSenEnc$, {\RAKE} keywords extractor, and knowledge base ${\KNOWBASE}$ \\
\textbf{Output:} Pseudo Label of $R^i$
\begin{algorithmic}[1]
\State Extract keywords $K^i$ from $s^i$ using {\RAKE}
\State Find $\bar{R}=\{T(t) | \KNOWTRIP \in \KNOWBASE$ and $\exists \KEYWORD^i_j \in K^i$, s.t. $\KEYWORD^i_j \in \KNOWTRIP\}$
\State Encode each $\bar{r}_j \in \bar{R}$ to $\UniSenEnc_j^r$ using {\UniSenEncFull}
\State Encode $[s_{i-1}, s_i]$ to $\UniSenEnc^s$
\State Compute cosine similarity $score$ between each $\UniSenEnc_j^r$ and $\UniSenEnc^s$
\State \textbf{return} $\bar{r}_j$s with the top $N$ highest $score$ 
\end{algorithmic}
}
\end{algorithm}

\subsection{Building Pseudo Label of $R^i$}
\label{sec:pseduolabel}

The main challenge for controlling generation with knowledge is that we have no explicit access to the hidden, latent controlling knowledge humans use to supervise their story writing. That means $R^{i}$, the knowledge associated with $s^{i}$ is not available. We, therefore, propose to use a weakly supervised signal to build the pseudo labels of $R^i$ from $s^i$. We hypothesize that $R^i$ should 1) overlap with $s^{i}$ in terms of keywords and 2) have strong connections to both the preceding sentence $s^{i-1}$ and $s^i$.  \textcolor{black}{We include $s^{i-1}$ along with $s^i$ because it is hard to retrieve appropriate knowledge using only $s^{i}$ due to the ambiguity of natural language. We also did not include other previous context beyond $s^{i-1}$ as additional context overwhelms the information contained in $s^{i}$.}


Following our hypothesis, we first extract keywords $K^i$ from $s^i$ using {\RAKE} \cite{rose2010automatic} and then match $K^i$ with all knowledge triples in ${\KNOWBASE}$. Transforming the retrieved triples into knowledge sentences gives us our set of $\Bar{R}^i$. We then take the sentence $s^i$ and $s^{i-1}$, concatenate them, and encode them using the Universal Sentence Encoder ({\UniSenEncFull}) \cite{cer2018universal}, a widely-used toolkit for semantic similarity, $U^s=U([s^{i-1}, s^i])$,  where we denote the encoder of {\UniSenEncFull} as $\UniSenEnc$. 
For each $\bar{r}_j^i\in\bar{R}^i$, we then calculate the cosine similarity between $U^s$ and $U^r_j=U(\bar{r}_j)$ and sort $\bar{R}^i$ based on this score. We take the top $N$ highest scores $\bar{r}_j^i$ as a pseudo label of $R^i$. Algorithm \ref{alg:kb_retrieval} describes this process. During the training phase of each  following model, we use this pseudo label of $R^i$ to represent $R^i$.

\subsection{Contextual Knowledge Ranker}
\label{sec:contextual_ranker}
While knowledge retrieval with keywords reduces the controlling knowledge candidate space from the knowledge base $\KNOWBASE$ to the subset $\hat{R}^i$, this set is still large and noisy since words are ambiguous and can have multiple senses. We, therefore, contextualize the knowledge sentences in $\hat{R}^i$ to obtain relevant and useful ones under $X^{i-1}$. To do this, we develop a contextual knowledge ranker.
The model is trained with pseudo-labels extracted with access to the future sentence $s^i$ as described in Sec. \ref{sec:pseduolabel}.

We use a {\BERT} model to encode both the context $X^{i-1}$ and each knowledge sentence $\hat{r}_j^i\in\hat{R}^i$. To adapt to the format of {\BERT}, we append a [SEP] token to each $R^j$ and $s^j$ inside the context $X^{i-1}$. A [CLS] token is then added to the beginning of $X^{i-1}$. For segment ids, we mark the tokens from the knowledge base as 0 and those from the story as 1. The representation of $X^{i-1}$ and $\hat{r}^i_j$  are then obtained after applying a linear layer on top of the embedding of the [CLS] token:
\begin{align*}
    V_x &= W_1 \, \text{\BERT}_{\text{CLS}}(X^{i-1}), \\
    V_j &= W_2 \, \text{\BERT}_{\text{CLS}}(\hat{r}_j^i), 
\end{align*}
where $W_1$ and $W_2$ are learnable weights. We then calculate the relevance {\it score} $C$ between  $X^{i-1}$ and $\hat{r}^i_j$ using the inner product between $V_x$ and $V_j$ as :
\begin{equation}
        C_j^i = C(X^{i-1}, \hat{r}^i_j) = V_x  V_j
\end{equation}
We take $R^i$ (Sec. \ref{sec:pseduolabel}) as positive samples and $\hat{R}^i \backslash R^i$ as negative samples to train our ranker.
Given a positive and a negative knowledge sentence $r_p$ and $r_n$, we define the ranking loss as
{\small 
\begin{equation}
\label{eq:loss}
       L = \max\{0, M-C(X^{i-1}, r_p) + C(X^{i-1}, r_n)\} 
\end{equation}}
where $M$ is a margin and determined empirically. Algorithms \ref{alg:knowledge_ranking} describe the ranker training process.  

At inference time, we simply calculate $C_j$ for all $\hat{r}_j^i\in \hat{R}^i$, sort them based on $C_j^i$ score and pick the top $N$ most relevant knowledge sentences as $R^i$.

\begin{algorithm}[h]
{\fontsize{10.1pt}{10.1pt}\selectfont
\caption{Knowledge Ranker Training}
\label{alg:knowledge_ranking}
\textbf{Parameters:} {\BERT} model parameters $\Theta$ and ranker model parameters $W_1$ and $W_2$ \\
\textbf{Input:} A story $S^l$ with $l$ sentences and a knowledge base ${\KNOWBASE}$
\begin{algorithmic}[1]
\State Initialize $\Theta$ using a pre-trained {\BERT} model and $W_1, W_2$ randomly. 
\State Dataset $D = \emptyset$
\State Call Algorithm \ref{alg:kb_retrieval} to retrieve $\RetKnowTrip^{1}$ from ${\KNOWBASE}$ using $s^1$.
\For{$i \in \{2, \ldots, l\}$ }
  \State Call Algorithm \ref{alg:kb_retrieval} to retrieve $\RetKnowTrip^{i}$ using $s^i$.
  \State Get $\hat{R}^i$ using knowledge retrieval (Section \ref{sec:know-ret})
  \For{$j \in \{1, \ldots, N\}$ }
      \State Sample $r_p$ from $R^i$ and $r_n$ from  $\hat{R}^i \backslash R^i$
      \State $D = D \cup (X^{i-1}, r_p, r_n)$
  \EndFor
\EndFor
\For{ $(X, r_p, r_n) \in D$ }
    \State Calculate loss $L$ using Equation \ref{eq:loss}
    \State Optimize $\text{\BERT}, W_1, W_2$
\EndFor
\State \textbf{return} $\text{\BERT}, W_1, W_2$
\end{algorithmic}
}
\end{algorithm}

\subsection{Conditional Generator}
\label{sec:cond-gen}

The conditional generator is a language model that incorporates the controlling knowledge and generates the following sentences. It concatenates the story context $X^{i-1}$ and controlling knowledge $R^i$ as input and generates $s^i$. A {\GPTTWO} transformer is used to model this conditional probability distribution. We describe the concatenated input representation in the Appendix \ref{subsec:input-representations}.

\section{Experimental Setup}

\subsection{Datasets}
\label{subsec:datasets}
We use the {\ROC} story dataset \cite{mostafazadeh2016story} for our experiments. It consists of 98,161 stories, where each story contains five sentences. 88,344/4,908/4,909 stories are used for train/validation/test sets, respectively. Following \citet{guan2020knowledge}, for each sentence, delexicalization is performed by replacing all the names and entities in stories with special placeholders, [{\it MALE}],
[{\it FEMALE}], and [{\it NEUTRAL}] for male, female
and unknown names and entities, respectively. Given the first sentence of each story, our model's task is to generate the rest of the story.  For our external knowledge base, we use ConceptNet \cite{speer2012representing}, consists of 600k knowledge triples.  

\subsection{Models}

We used Megatron-LM \cite{Megatron} for pre-trained {\BERT} and {\GPTTWO} models to initialize our contextual knowledge ranker and generative models, respectively.  For the model configurations, hidden size, number of layers, and attention heads, we used the configurations of {\BERT} and {\GPTTWO} as in Megatron-LM. For generation with our {\GPTTWO} models, we used a top-$k$ sampling scheme \cite{fan2018hierarchical} with $k=40$ and a softmax temperature of 0.7. We detail the training hyperparameters and the input representations for {\GPTTWO} and {\BERT} in Appendix \ref{subsec:gpt2-hyperparams} \& \ref{subsec:bert-hyperparams} . Both the keyword \textcolor{black}{predictor} and the conditional sentence generator follow the same settings.

To train our contextual knowledge ranker, we set the margin to 5.0. We set the number of knowledge sentences in $R^i$ to 10. Therefore, for a given story context, the top 10 retrieved knowledge sentences from ConceptNet according to {\UniSenEncFull} are chosen as the positive samples. We further select 40 negative samples to compute our margin loss. 
We then randomly sample 50 (positive, negative) pairs for each story context to train our contextual knowledge ranker. In total, we used $\sim$15 million
pairs for training and 
$\sim$1 million pairs for validation. After training our ranker, we achieve a validation accuracy of 0.9.

\subsection{Controllability Experiment Setup}
\label{sec:controllability_setup}
To test the controllability of our model, we perform experiments where we change keywords to their antonyms. With antonyms, we expect maximal change to the story generation. To do that, we first use {\OurIIVIIm} to generate keywords $K$ and corresponding full story $S$. Then we identify the first keyword $k^i_a\in K^i$ from $K$ whose antonym is available at WordNet \cite{miller1995wordnet}. If multiple antonyms for $k^i_a$ are available we sample one with a uniform random probability. Afterwards, we provide the start of story $\{s^1, s^2, \cdots, s^{i-1}\}$, the keywords shared with our original story $\{K^1, K^2, \cdots, K^{i-1}\}$, and the antonym of $k^i_a$ to either {\OurIIVIIm} or larger models (e.g. {\OurIIIIVVm}). We then let the model finish the generation. We refer to these generations as $\OurANTONYM$, for example, we call the antonym generations from {\OurIIIIVVm} model as \OurIIIIVVm-\ANTONYM.

\subsection{Baselines}
We compare our model with the following state-of-the-art story generation models.
(1) \textbf{Plan and write \cite{yao2019plan}:} The authors use an LSTM-based model to first generate a sequence of keywords for planning the story. These keywords are then used to condition the generation.  (2) \textbf{Knowledge enhanced {\GPTTWO}  \cite{guan2020knowledge}:} This work is currently the SOTA for ROC story generation. It finetunes a pre-trained {\GPTTWO}  model with knowledge triples from commonsense datasets. Similar to our method, the knowledge triples are converted to sentences with templates. A multitask learning framework is then developed to further finetune the story generation task and classify corrupted stories from real ones. 
We do not compare to \citet{fan2019strategies} because \citet{guan2020knowledge} has already shown their model significantly outperforms \citet{fan2019strategies} and in this work, we compare to \citet{guan2020knowledge}.
(3) \textbf{$\GPTTWO$-124M:} This baseline finetunes a {\GPTTWO} model with a next token prediction loss on the story. 

\subsection{Evaluation}
\label{sec:eval_setup}
We conduct both automatic as well as human evaluations to assess our generation. 

\subsubsection{Automatic Evaluation}
We use the following metrics to compare different models: 
\textbf{Repeat:} measures the redundancy of the generated story by reporting the percentage of the stories that contain at least one repeated 4-gram \cite{shao2019long}.
\textbf{Distinct:} measures the diversity of generated stories by reporting the ratio between distinct 4-grams to all generated 4-grams. \textbf{Perplexity:} In the inference phase, our models involve two steps of generation: (i) generate set of knowledge sentences, $R^i$ from story context $X^{i-1}$, (ii) generate story sentence, $s^i$ from $X^{i-1}$ and $R^i$. To report the perplexity of the conditional generator we sample $R^i$ sequentially before generating each story sentence $s^{i}$ and report the total perplexity of all sentences $s^i$ for $i\in[2,l]$ where $l$ is the number of sentences in the story.

\subsubsection{Human Evaluation on Quality}
We conduct human evaluations on Amazon Mechanical Turk\footnote{https://www.mturk.com/} (AMT) to analyze the quality of our generations on three aspects: {\bf Fluency}, {\bf Coherence}, and {\bf Consistency}. To evaluate fluency, we show the annotators a pair of generated stories from two models. We ask them to evaluate each sentence independently and choose the story with better overall fluency. Fluency of a story is defined as a measure of intra-sentence linguistic quality and grammatical correctness taken over all sentences of the story. For coherence, we provide the same stories as in fluency
but ask to choose the one with better inter-sentence causal and temporal dependencies. We let the annotators choose {\it tie} for both fluency and coherence. 

Different from the settings of fluency and coherence, we only show one generated story to annotators to evaluate consistency. They are required to choose whether the story is logically consistent, based on whether the story self contradicts or not. 

We set up these three evaluations as independent AMT tasks to make sure the tasks do not influence each other and introduce spurious correlations between labels. To reduce noise in our labeling process, we only accepted workers with an approval rating over 90\% and have over 1k accepted jobs. We further limited the location of the annotators to the United States. For each example, we explicitly ask them to spend at least 15 seconds to evaluate coherency and 10 seconds to evaluate the other two properties. 
In total, we randomly sample 200 stories and assign five annotators for each story. We adopted majority voting to make final decisions among the five annotators.

\subsubsection{Human Evaluation on Controllability}
To evaluate how controllable our model is, we conduct another human evaluation just for controllability. We show the annotators the start of a story, original keywords, and the corresponding generation. We then show the antonyms of the keywords, along with the corresponding generated story, and ask the annotators if the new story has changed compared to the original story in accordance with the meaning of the keyword's antonyms. The rest of the AMT settings for these experiments are the same as our consistency experiments.

\section{Results}

In this section, we first perform automatic and human evaluations with different model sizes and compare our framework to the existing baselines. We then evaluate the controllability of our model and finally show ablation study varying {\GPTTWO} and {\BERT} model sizes. The detailed configuration of the model sizes are shown in Table \ref{table:configure}. We provide several generated stories in Appendix \ref{subsec:generation-examples} varying the length of the given context. \textcolor{black}{We use {\MODEL} to denote {\OURMODEL} in the tables due to the limited space.}

\begin{table}[!htb]
\centering\scalebox{0.8}{
\begin{tabular}{c@{\hskip3pt}cccc}
\hline
 & Conditional & Keyword  &  Knowledge  \\
Model Name\ \ \ \  & Generator & Generator &  Ranker  \\
 & (\GPTTWO) & (\GPTTWO) &  (\BERT)  \\
\hline
\IIVIIm & 124M	& 124M	& 336M		\\
\IIIIVVm & 355M	& 355M	& 336M		 \\
\VIIVIVIIIm & 774M   & 774M	& 336M		\\
\IIb   & 2.5B     & 2.5B	& 336M	\\
\VIIIb 	& 8.3B	& 2.5B	& 336M \\
\hline
\end{tabular}
}
\vspace{-2mm}
\caption{\label{table:configure} Number of parameters of our models ({\MODEL} is the short form for {\OURMODEL}).}
\vspace{-2mm}
\end{table}

\subsection{Automatic and Human Evaluations}

\begin{table*}[!htb]
\centering\scalebox{0.85}{
\begin{tabular}{lcc@{\hskip3pt}r}
\hline 
Source A & Coherence $\uparrow$ & Fluency $\uparrow$  & Source B \\
\hline\hline
{ \IIVIIm} &  {\textbf {78.5\%}} - 13.0\%	& {\textbf {66.5\%}} - 22.5\% & { {\citet{YaoPlanAndWrite2018}}} \\
{ \IIVIIm} &  {\textbf {46.0}}\% - 39.0\%	& {\textbf {44.5}}\% - 43.5\% 	 & { \citet{guan2020knowledge}} \\
{ \IIIIVVm} &  {\textbf {56.0}}\% - 30.5\%	& {\textbf {46.5}}\% - 30.5\%	 & { \citet{guan2020knowledge}} \\ \hline
{ \IIIIVVm} &  {\textbf {52.0}}\% - 31.5\%	& {\textbf {46.5}}\% - 39.0\%	& { \IIVIIm} \\
{ \VIIVIVIIIm} &  {\textbf {44.5}}\% - 41.5\%	& {\textbf {56.0}}\% - 33.5\%	& { \IIIIVVm} \\
{ \IIb} &  {\textbf {50.5}}\% - 30.5\%	& {\textbf {53.0}}\% - 39.0\%	& { \VIIVIVIIIm} \\
{ \VIIIb} &  \textbf {46.0}\% - 39.5\%	& 46.5\% - {46.5}\%	& { \IIb} \\
\hline
\end{tabular}
}
\caption{\label{table:all-allgorithms-human-evaluations} 
Pairwise comparison between our models and baselines. Percentages in the format “A\% - B\%” indicate how often annotators rank the samples from source A  better than from source B for a given category, and
vice versa. Percentage pairs do not sum to 100\%
as the annotators were allowed to choose ``tie" as being of equal quality.
$\OurIIVIIm$ achieves better results than all baselines. Scaling the models shows better coherence and fluency. }
\vspace{-3mm}
\end{table*}

\begin{table}[!htb]
\centering\scalebox{0.72}{
\begin{tabular}{l@{\hskip3pt}ccc|c}
\hline 
\multirow{2}{*}{Name} & \multirow{2}{*}{PPL $\downarrow$}   & \multirow{2}{*}{Repeat $\downarrow$} & \multirow{2}{*}{Distinct $\uparrow$} & Consistency $\uparrow$ \\
 & & & & (Human Eval)  \\
\hline\hline
{\GPTTWO-124M} & 6.98		& 27.2	& 74.1 & 69.5 \\
\hline
{\small \citet{YaoPlanAndWrite2018}} & NA		& {\textbf{13.3}}	& 63.7 & 49.0 \\
{\small \citet{guan2020knowledge}} & 7.04 	& 22.1	& 77.1 &  67.0\\
\hline
\IIVIIm & 9.37	& 20.0	& 80.1 & 74.5 \\
\IIIIVVm & 8.02		& 19.9	& 81.6 & 75.5\\
\VIIVIVIIIm & 6.58    	& 21.3	& 81.6 & 80.5\\
\IIb   & 6.31    	& 21.2	& 82.6 & 89.0\\
\VIIIb   & {\textbf{6.21}}	& 21.2	& {\textbf{82.8}}  & {\textbf{93.0}}\\
\hline
\end{tabular}
}
\caption{\label{table:all-allgorithms-baseline-metrics} Evaluation results for the previous state-of-the-art models as well as our algorithm at different sizes. Perplexity, repeat, and distinct are evaluated automatically whereas consistency is obtained using human evaluations. Our smallest model with 124M parameters achieves better distinct and consistency score compared to prior work. Increasing model size up to 8B improves perplexity, distinct, and consistency scores. For reference, the ground truth human writing gives 7.6 score for repeat and 88.9 for distinct.}
\end{table}

Table \ref{table:all-allgorithms-baseline-metrics} shows that our smallest model, {\OurIIVIIm} achieves better distinct and consistency scores compared to previous work. For repetition, our model is worse than \citet{yao2019plan} which was also observed in \citet{guan2020knowledge}. The reason could be their small 8M  model is better at learning short term statistics (e.g. 4-grams), while large models are better at learning long term dependencies. Compared to other {\GPTTWO} based models (i.e. {\GPTTWO-124M} and \citet{guan2020knowledge}), {\OurIIVIIm} achieves lower repeat and higher distinct scores, hence our model generates less repetitive stories.

We notice from Table \ref{table:all-allgorithms-baseline-metrics} that our perplexity (PPL) score is much higher than other \GPTTWO-based models. Our hypothesis for why this occurs is that other {\GPTTWO}-based methods directly model and report $P(s^i|s^1, s^2, \cdots, s^{i-1})$ while our conditional generator models and reports $P(s^i| X^{i-1}, R^i)$. When computing perplexity, $[s^1, s^2, \cdots, s^{i-1}]$ are given ground truth tokens, but $R^i$ and all $R$ in $X^{i-1}$ must be sampled from a distribution that is learned with weak supervision. This sampling introduces 
noise and non-determinism that results in higher perplexity. This discrepancy is not an issue when analyzing automatic evaluation metrics within our model family. When scaling our model from 124M up to 8B parameters we see a consistent drop in PPL and increase in distinct. This shows larger models can generate better stories with more diversity.


Human evaluation results are presented in last column of Table \ref{table:all-allgorithms-baseline-metrics} (consistency) and in Table \ref{table:all-allgorithms-human-evaluations}. Comparing {\OurIIVIIm} to \citet{yao2019plan}, we achieve much better coherence, fluency, and consistency scores, which shows the benefit of large pre-trained transformer models. Comparing {\OurIIVIIm} to \citet{guan2020knowledge} which uses a similar base model, we find that fluency is similar, however we should note that \citet{guan2020knowledge} is not controllable and our model has significantly better coherence (+7.0\%) in Table \ref{table:all-allgorithms-human-evaluations} and consistency (+7.5\%) in Table \ref{table:all-allgorithms-baseline-metrics}. We attribute this to the use of the retrieved knowledge, $R^i$. By explicitly providing facts pertinent to the next sentence, the conditional generative model can focus on just generating text. By comparison, a standard autoregressive {\GPTTWO} model is tasked with predicting both the topics and the text of the next sentence.   

Scaling this up, and comparing {\OurIIIIVVm} to \citet{guan2020knowledge}, our model 
significantly outperforms in all aspects. Furthermore, a thorough comparison among {\OurIIIIVVm}, {\OurVIIVIVIIIm}, {\OurIIb}, {\OurVIIIb} shows that scaling the model size further almost always improves the quality of generation in terms of fluency, coherence and consistency.  For consistency, our best model at 8B parameters achieves a score of 93\%.  

\subsection{Controllability Evaluation}

We evaluate the controllability by changing keywords to their antonyms as detailed in Section \ref{sec:controllability_setup} \& \ref{sec:eval_setup}. Table \ref{table:all-allgorithms-control-metrics} shows repeat and distinct for {\OurIIVIIm} as well as the controlled versions at three different sizes. Altering control with antonym keywords gives lower repeat and higher distinct scores than the original generation. As the model size increases, the repeat stays almost constant while distinct improves. These results show that changing keywords manually results in distinct and not repeated text.  


\begin{table}[!htb]
\centering\scalebox{0.8}{
\begin{tabular}{l@{\hskip3pt}cccccc}
\hline 
Name & Repeat $\downarrow$ & Distinct $\uparrow$  \\
\hline\hline

\hline

{\IIVIIm } 		& 20.0	& 80.1  \\ \hline
{\IIVIIm-\ANTONYM}  & 17.8	& 80.9 \\
{\IIIIVVm-\ANTONYM}   	& 18.0	& 81.6 \\
{\VIIIb-\ANTONYM} & 18.5	& 82.8   \\
\hline
\end{tabular}
}
\caption{\label{table:all-allgorithms-control-metrics} Comparing controllability of the models by changing the keywords to their antonyms. Controlled generations show less repetition and higher diversity compared to the original one.}
\end{table}


Further supporting this hypothesis, evaluation of controllability in Table \ref{table:all-allgorithms-human-evaluations-control} shows that {\OurIIVIIm-\ANTONYM} achieves a high controllability score of 77.5\%. This means that by changing the keywords to their antonym, 77.5\% of newly generated stories change their semantic content to follow the new antonym keywords. We also show that larger models are better able to leverage keyword control. Scaling the model size from 124M to 355M and 8B model further improves the controllability score to 84.5\% 
and 91.5\%, respectively. We again observe the quality (e.g. coherence) of our controlled generation improves as the model size scales to 8B.

\begin{table}[!htb]
\centering\scalebox{0.8}{
\begin{tabular}{l@{\hskip3pt}c@{\hskip3pt}r}
\hline 
Name & Controllability $\uparrow$ \\
\hline\hline
{\IIVIIm-\ANTONYM} 	 & 77.5\%\\
{\IIIIVVm-\ANTONYM}	& 84.5\% \\
{\VIIIb-\ANTONYM}	& \textbf{91.5}\% \\
%
\hline
\end{tabular}
}
\caption{\label{table:all-allgorithms-human-evaluations-control} 
Human evaluation for controllability by changing keywords to their antonyms. Over 77\% of our generation changes according to the keywords.}
\vspace{-3mm}
\end{table}

\subsection{Ablation Studies}
In this section, we conduct the ablation study on the planning strategy and external knowledge. The study of model size can be found in the Appendix \ref{subsec:ablation_model_size}.
\subsubsection{Planning Strategy}
\begin{table}[t]

\centering\scalebox{0.8}{
\begin{tabular}{l@{\hskip3pt}c@{\hskip3pt}c@{\hskip3pt}c}
\hline 
Name & Repeat $\downarrow$ & Distinct $\uparrow$ \\
\hline\hline
{\IIVIIm} (D)            & 20.04	& 80.14	\\
{\IIVIIm}  w/o knowledge (D)	& 23.59	& 79.39  \\
{\IIVIIm}  (S)		& 23.87	& 79.45  \\
{\IIVIIm}  w/o knowledge (S)  & 23.98	& 79.61  \\

\hline
\end{tabular}
}
\caption{\label{table:ablation_strategy} \textcolor{black}{Ablation studies of static (S) vs dynamic (D) planning strategy, with and without knowledge.}}
\vspace{-3mm}
\end{table}
\textcolor{black}{
In this section, we investigate the effects of planning strategy in our framework. \citet{yao2019plan} showed that static planning works better than dynamic planning for LSTM-based models. To introduce the static planning in our model, we predicted all the keywords and relevant knowledge sentences from the starting sentence and generated the entire stories. When we compare these generations with the stories generated by dynamic planning, we see in Table \ref{table:ablation_strategy} (first and third rows) that dynamic planning outperforms the static planning strategy with higher distinction (+0.7\%) and lower repetition (-3.8\%) scores. This is due to direct guidance over each sentence provided by the retrieved knowledge from dynamic planning . In contrast, in static planning, the retrieved knowledge sentences are all predicted together at the beginning using only the starting sentence, which makes the supervision for each story sentence weaker and noisier. }

\subsubsection{External Knowledge}
\textcolor{black}{
In this section, we investigate the importance of retrieved knowledge. Table \ref{table:ablation_strategy} (first and second rows) shows that, when excluding the knowledge from our framework (i.e. {\OurIIVIIm} w/o knowledge), distinction score decreases by 0.8\% and repetition increases by 3.6\%, highlighting the importance of external knowledge in our approach. Unlike dynamic planning, we observe that in static planning, the external knowledge does not play an important role in the quality of the generations and using or not using the knowledge leads to similar qualities. This observation also confirms that knowledge needs to be planned dynamically.}

\section{Future Work}
\textcolor{black}{
Short story sentences in {\ROC} story dataset limits our exploration from several potential research directions.
For example, how long the control signal would propagate for longer generations? Investigating this issue using longer story datasets (e.g. {\sc{WritingPrompts}} \cite{fan2018hierarchical}) is a subject for future work. Other interesting direction may include incorporating the structure-level controllability by adding it as either an extra input for the conditional generator or a multitask learning supervision for each sequence.}

\textcolor{black}{We also observed that in some cases during the generation, our model simply mentions the given word in the sentence, and talks about things that are not strictly related to or restricted by the given word. For example, the generated story of {\OurVIIIb} in Table \ref{tab: example_15} only mentions the keyword ``realize" instead of centering around it. This is caused by the RAKE keywords extractor, which does not always extract the keywords that represent the sentence well. One way to mitigate this issue is to leverage longer context information to identify better keywords which is subject of the future work.
}

\section{Related Work}

\paragraph{Knowledge} Incorporation of knowledge into language models has shown promising results for downstream tasks, such as factual correct generation \cite{logan2019barack} ,
commonsense knowledge graph construction \cite{bosselut2019comet}, 
entity typing \cite{zhang2019ernie} and etc. 
More recently, several works have shown that inclusion of learned mechanisms for explicit or implicit knowledge can lead to the state-of-the-art results in Question Answering \cite{REALM,DPR,ORQA,RAGS} and dialogue modeling \cite{Blendr}.

\paragraph{Storytelling} There are several different storytelling tasks described throughout the literature. Storytelling can be classified into story completion ~\cite{chen2019incorporating}, story ending generation ~\cite{guan2019story}, story generation from prompts~\cite{fan2018hierarchical} or titles ~\cite{yao2019plan}, and story generation from a given sentence ~\cite{guan2020knowledge}. Different approaches have been developed to model the structure of stories with storylines ~\cite{yao2019plan}, skeletons ~\cite{xu2018skeleton}, Conditional Variational AutoEncoders ~\cite{wang2019t} and a coarse-to-fine framework ~\cite{fan2019strategies}. Other works focus on incorporating commonsense knowledge into story generation with attention-based models \cite{guan2019story, chen2019incorporating}. Recently, pre-trained language models have been used to finetune on both story completion datasets and commonsense knowledge to further improve the quality of story completion \cite{guan2020knowledge}. However, few works concern the controllability of language model generation, especially for the large pre-trained models that are common in today's literature. 

\paragraph{Controllable Generation} Controllable text generation has a wide range of applications, including controlling through persona \cite{zhang2018personalizing,boyd2020large}, politeness \cite{niu2018polite}, etc. \textcolor{black}{\citet{wiseman2018learning} presented controlling generations by learning  latent, discrete templates from data. \citet{fu2019rethinking} discovered the importance of pivot words that determines the sentence attributes and presented a lexical analysis framework. To control large pre-trained models,} \citet{keskar2019ctrl} demonstrated the ability to control text generation through a wide range of aspects, such as domains and links. 
Plug-and-play language models ~\citet{dathathri2019plug} also address whole document controllability by adding a linear classifier on top of {\GPTTWO} to predict whether generated text observes a particular style or property. 
\citet{prabhumoye2020exploring} provides a good survey of five modules for control. Differing from these works, we control the generation through keywords backed by external knowledge.

\section{Conclusion}
In this paper, we proposed a novel framework that adds control to text generation with external knowledge. Our model first generates a set of keywords and a knowledge retriever then queries an external knowledge base for triples related to the keywords. Based on the relevance to the story context, a contextual knowledge ranker ranks the retrieved knowledge sentences and feeds the top ones to a conditional generator to generate the next story sentence. Experimental results on the {\ROC} story dataset showed that our model outperforms state-of-the-art models by generating less repetitive, more diverse and  logically consistent stories. 
Human evaluation of the controllability of our model shows that 91.5\% of generated stories are successfully controlled by changing keywords to their antonym. 
In line with current trends, we also demonstrate that using larger pre-trained language models consistently improves both the quality of the generated stories and controllability. 


\bibliography{anthology,emnlp2020}

\begin{thebibliography}{41}
\expandafter\ifx\csname natexlab\endcsname\relax\def\natexlab#1{#1}\fi

\bibitem[{Bosselut et~al.(2019)Bosselut, Rashkin, Sap, Malaviya, Celikyilmaz,
  and Choi}]{bosselut2019comet}
Antoine Bosselut, Hannah Rashkin, Maarten Sap, Chaitanya Malaviya, Asli
  Celikyilmaz, and Yejin Choi. 2019.
\newblock Comet: Commonsense transformers for automatic knowledge graph
  construction.
\newblock In \emph{Proceedings of the 57th Annual Meeting of the Association
  for Computational Linguistics}, pages 4762--4779.

\bibitem[{Boyd et~al.(2020)Boyd, Puri, Shoeybi, Patwary, and
  Catanzaro}]{boyd2020large}
Alex Boyd, Raul Puri, Mohammad Shoeybi, Mostofa Patwary, and Bryan Catanzaro.
  2020.
\newblock Large scale multi-actor generative dialog modeling.
\newblock \emph{arXiv preprint arXiv:2005.06114}.

\bibitem[{Cer et~al.(2018)Cer, Yang, Kong, Hua, Limtiaco, John, Constant,
  Guajardo-Cespedes, Yuan, Tar et~al.}]{cer2018universal}
Daniel Cer, Yinfei Yang, Sheng-yi Kong, Nan Hua, Nicole Limtiaco, Rhomni~St
  John, Noah Constant, Mario Guajardo-Cespedes, Steve Yuan, Chris Tar, et~al.
  2018.
\newblock Universal sentence encoder.
\newblock \emph{arXiv preprint arXiv:1803.11175}.

\bibitem[{Chen et~al.(2019)Chen, Chen, and Yu}]{chen2019incorporating}
Jiaao Chen, Jianshu Chen, and Zhou Yu. 2019.
\newblock Incorporating structured commonsense knowledge in story completion.
\newblock In \emph{Proceedings of the AAAI Conference on Artificial
  Intelligence}, volume~33, pages 6244--6251.

\bibitem[{Dathathri et~al.(2019)Dathathri, Madotto, Lan, Hung, Frank, Molino,
  Yosinski, and Liu}]{dathathri2019plug}
Sumanth Dathathri, Andrea Madotto, Janice Lan, Jane Hung, Eric Frank, Piero
  Molino, Jason Yosinski, and Rosanne Liu. 2019.
\newblock Plug and play language models: a simple approach to controlled text
  generation.
\newblock \emph{arXiv preprint arXiv:1912.02164}.

\bibitem[{Devlin et~al.(2018)Devlin, Chang, Lee, and Toutanova}]{bert}
Jacob Devlin, Ming-Wei Chang, Kenton Lee, and Kristina Toutanova. 2018.
\newblock Bert: Pre-training of deep bidirectional transformers for language
  understanding.
\newblock \emph{arXiv preprint arXiv:1810.04805}.

\bibitem[{Fan et~al.(2018)Fan, Lewis, and Dauphin}]{fan2018hierarchical}
Angela Fan, Mike Lewis, and Yann Dauphin. 2018.
\newblock Hierarchical neural story generation.
\newblock In \emph{Proceedings of the 56th Annual Meeting of the Association
  for Computational Linguistics (Volume 1: Long Papers)}, pages 889--898.

\bibitem[{Fan et~al.(2019)Fan, Lewis, and Dauphin}]{fan2019strategies}
Angela Fan, Mike Lewis, and Yann Dauphin. 2019.
\newblock Strategies for structuring story generation.
\newblock In \emph{Proceedings of the 57th Annual Meeting of the Association
  for Computational Linguistics}, pages 2650--2660.

\bibitem[{Fu et~al.(2019)Fu, Zhou, Chen, and Li}]{fu2019rethinking}
Yao Fu, Hao Zhou, Jiaze Chen, and Lei Li. 2019.
\newblock Rethinking text attribute transfer: A lexical analysis.
\newblock In \emph{Proceedings of the 12th International Conference on Natural
  Language Generation}, pages 24--33.

\bibitem[{Gokaslan and Cohen(2019)}]{openwebtext}
Aaron Gokaslan and Vanya Cohen. 2019.
\newblock Openwebtext corpus.

\bibitem[{Guan et~al.(2020)Guan, Huang, Zhao, Zhu, and
  Huang}]{guan2020knowledge}
Jian Guan, Fei Huang, Zhihao Zhao, Xiaoyan Zhu, and Minlie Huang. 2020.
\newblock A knowledge-enhanced pretraining model for commonsense story
  generation.
\newblock \emph{arXiv preprint arXiv:2001.05139}.

\bibitem[{Guan et~al.(2019)Guan, Wang, and Huang}]{guan2019story}
Jian Guan, Yansen Wang, and Minlie Huang. 2019.
\newblock Story ending generation with incremental encoding and commonsense
  knowledge.
\newblock In \emph{Proceedings of the AAAI Conference on Artificial
  Intelligence}, volume~33, pages 6473--6480.

\bibitem[{Guu et~al.(2020)Guu, Lee, Tung, Pasupat, and Chang}]{REALM}
Kelvin Guu, Kenton Lee, Zora Tung, Panupong Pasupat, and Ming-Wei Chang. 2020.
\newblock Realm: Retrieval-augmented language model pre-training.
\newblock \emph{arXiv preprint arXiv:2002.08909}.

\bibitem[{Karpukhin et~al.(2020)Karpukhin, O{\u{g}}uz, Min, Wu, Edunov, Chen,
  and Yih}]{DPR}
Vladimir Karpukhin, Barlas O{\u{g}}uz, Sewon Min, Ledell Wu, Sergey Edunov,
  Danqi Chen, and Wen-tau Yih. 2020.
\newblock Dense passage retrieval for open-domain question answering.
\newblock \emph{arXiv preprint arXiv:2004.04906}.

\bibitem[{Keskar et~al.(2019)Keskar, McCann, Varshney, Xiong, and
  Socher}]{keskar2019ctrl}
Nitish~Shirish Keskar, Bryan McCann, Lav~R. Varshney, Caiming Xiong, and
  Richard Socher. 2019.
\newblock \href {http://arxiv.org/abs/1909.05858} {Ctrl: A conditional
  transformer language model for controllable generation}.

\bibitem[{Kingma and Ba(2014)}]{kingma2014adam}
Diederik~P Kingma and Jimmy Ba. 2014.
\newblock Adam: A method for stochastic optimization.
\newblock \emph{arXiv preprint arXiv:1412.6980}.

\bibitem[{Lee et~al.(2019)Lee, Chang, and Toutanova}]{ORQA}
Kenton Lee, Ming-Wei Chang, and Kristina Toutanova. 2019.
\newblock Latent retrieval for weakly supervised open domain question
  answering.
\newblock \emph{arXiv preprint arXiv:1906.00300}.

\bibitem[{Lewis et~al.(2020)Lewis, Perez, Piktus, Petroni, Karpukhin, Goyal,
  K{\"u}ttler, Lewis, Yih, Rockt{\"a}schel et~al.}]{RAGS}
Patrick Lewis, Ethan Perez, Aleksandara Piktus, Fabio Petroni, Vladimir
  Karpukhin, Naman Goyal, Heinrich K{\"u}ttler, Mike Lewis, Wen-tau Yih, Tim
  Rockt{\"a}schel, et~al. 2020.
\newblock Retrieval-augmented generation for knowledge-intensive nlp tasks.
\newblock \emph{arXiv preprint arXiv:2005.11401}.

\bibitem[{Logan et~al.(2019)Logan, Liu, Peters, Gardner, and
  Singh}]{logan2019barack}
Robert Logan, Nelson~F Liu, Matthew~E Peters, Matt Gardner, and Sameer Singh.
  2019.
\newblock Barack’s wife hillary: Using knowledge graphs for fact-aware
  language modeling.
\newblock In \emph{Proceedings of the 57th Annual Meeting of the Association
  for Computational Linguistics}, pages 5962--5971.

\bibitem[{Miller(1995)}]{miller1995wordnet}
George~A Miller. 1995.
\newblock Wordnet: a lexical database for english.
\newblock \emph{Communications of the ACM}, 38(11):39--41.

\bibitem[{Mostafazadeh et~al.(2016)Mostafazadeh, Vanderwende, Yih, Kohli, and
  Allen}]{mostafazadeh2016story}
Nasrin Mostafazadeh, Lucy Vanderwende, Wen-tau Yih, Pushmeet Kohli, and James
  Allen. 2016.
\newblock Story cloze evaluator: Vector space representation evaluation by
  predicting what happens next.
\newblock In \emph{Proceedings of the 1st Workshop on Evaluating Vector-Space
  Representations for NLP}, pages 24--29.

\bibitem[{Niu and Bansal(2018)}]{niu2018polite}
Tong Niu and Mohit Bansal. 2018.
\newblock Polite dialogue generation without parallel data.
\newblock \emph{Transactions of the Association for Computational Linguistics},
  6:373--389.

\bibitem[{Prabhumoye et~al.(2020)Prabhumoye, Black, and
  Salakhutdinov}]{prabhumoye2020exploring}
Shrimai Prabhumoye, Alan~W Black, and Ruslan Salakhutdinov. 2020.
\newblock Exploring controllable text generation techniques.
\newblock \emph{arXiv preprint arXiv:2005.01822}.

\bibitem[{Radford et~al.(2018)Radford, Narasimhan, Salimans, and
  Sutskever}]{radford2018improving}
Alec Radford, Karthik Narasimhan, Tim Salimans, and Ilya Sutskever. 2018.
\newblock Improving language understanding by generative pre-training.

\bibitem[{Radford et~al.(2019)Radford, Wu, Child, Luan, Amodei, and
  Sutskever}]{radford2019language}
Alec Radford, Jeffrey Wu, Rewon Child, David Luan, Dario Amodei, and Ilya
  Sutskever. 2019.
\newblock Language models are unsupervised multitask learners.

\bibitem[{Roller et~al.(2020)Roller, Dinan, Goyal, Ju, Williamson, Liu, Xu,
  Ott, Shuster, Smith et~al.}]{Blendr}
Stephen Roller, Emily Dinan, Naman Goyal, Da~Ju, Mary Williamson, Yinhan Liu,
  Jing Xu, Myle Ott, Kurt Shuster, Eric~M Smith, et~al. 2020.
\newblock Recipes for building an open-domain chatbot.
\newblock \emph{arXiv preprint arXiv:2004.13637}.

\bibitem[{Rose et~al.(2010)Rose, Engel, Cramer, and Cowley}]{rose2010automatic}
Stuart Rose, Dave Engel, Nick Cramer, and Wendy Cowley. 2010.
\newblock Automatic keyword extraction from individual documents.
\newblock \emph{Text mining: applications and theory}, 1:1--20.

\bibitem[{Shao et~al.(2019)Shao, Huang, Wen, Xu et~al.}]{shao2019long}
Zhihong Shao, Minlie Huang, Jiangtao Wen, Wenfei Xu, et~al. 2019.
\newblock Long and diverse text generation with planning-based hierarchical
  variational model.
\newblock In \emph{Proceedings of the 2019 Conference on Empirical Methods in
  Natural Language Processing and the 9th International Joint Conference on
  Natural Language Processing (EMNLP-IJCNLP)}, pages 3248--3259.

\bibitem[{Shoeybi et~al.(2019)Shoeybi, Patwary, Puri, LeGresley, Casper, and
  Catanzaro}]{Megatron}
Mohammad Shoeybi, Mostofa Patwary, Raul Puri, Patrick LeGresley, Jared Casper,
  and Bryan Catanzaro. 2019.
\newblock Megatron-lm: Training multi-billion parameter language models using
  gpu model parallelism.
\newblock \emph{arXiv preprint arXiv:1909.08053}.

\bibitem[{Speer and Havasi(2012)}]{speer2012representing}
Robert Speer and Catherine Havasi. 2012.
\newblock Representing general relational knowledge in conceptnet 5.
\newblock In \emph{LREC}, pages 3679--3686.

\bibitem[{Trinh and Le(2018)}]{ccstories}
Trieu~H Trinh and Quoc~V Le. 2018.
\newblock A simple method for commonsense reasoning.
\newblock \emph{arXiv preprint arXiv:1806.02847}.

\bibitem[{Wang and Wan(2019)}]{wang2019t}
Tianming Wang and Xiaojun Wan. 2019.
\newblock T-cvae: Transformer-based conditioned variational autoencoder for
  story completion.
\newblock In \emph{Proceedings of the 28th International Joint Conference on
  Artificial Intelligence}, pages 5233--5239. AAAI Press.

\bibitem[{Wiseman et~al.(2018)Wiseman, Shieber, and Rush}]{wiseman2018learning}
Sam Wiseman, Stuart~M Shieber, and Alexander~M Rush. 2018.
\newblock Learning neural templates for text generation.
\newblock In \emph{Proceedings of the 2018 Conference on Empirical Methods in
  Natural Language Processing}, pages 3174--3187.

\bibitem[{Wolf et~al.(2019)Wolf, Sanh, Chaumond, and
  Delangue}]{wolf2019transfertransfo}
Thomas Wolf, Victor Sanh, Julien Chaumond, and Clement Delangue. 2019.
\newblock Transfertransfo: A transfer learning approach for neural network
  based conversational agents.
\newblock \emph{arXiv preprint arXiv:1901.08149}.

\bibitem[{Xu et~al.(2018)Xu, Ren, Zhang, Zeng, Cai, and Sun}]{xu2018skeleton}
Jingjing Xu, Xuancheng Ren, Yi~Zhang, Qi~Zeng, Xiaoyan Cai, and Xu~Sun. 2018.
\newblock A skeleton-based model for promoting coherence among sentences in
  narrative story generation.
\newblock In \emph{Proceedings of the 2018 Conference on Empirical Methods in
  Natural Language Processing}, pages 4306--4315.

\bibitem[{Yao et~al.(2019)Yao, Peng, Weischedel, Knight, Zhao, and
  Yan}]{yao2019plan}
Lili Yao, Nanyun Peng, Ralph Weischedel, Kevin Knight, Dongyan Zhao, and Rui
  Yan. 2019.
\newblock Plan-and-write: Towards better automatic storytelling.
\newblock In \emph{Proceedings of the AAAI Conference on Artificial
  Intelligence}, volume~33, pages 7378--7385.

\bibitem[{Yao et~al.(2018)Yao, Peng, Weischedel, Knight, Zhao, and
  Yan}]{YaoPlanAndWrite2018}
Lili Yao, Nanyun Peng, Ralph~M. Weischedel, Kevin Knight, Dongyan Zhao, and Rui
  Yan. 2018.
\newblock \href {http://arxiv.org/abs/1811.05701} {Plan-and-write: Towards
  better automatic storytelling}.
\newblock \emph{CoRR}, abs/1811.05701.

\bibitem[{Zellers et~al.(2019)Zellers, Holtzman, Rashkin, Bisk, Farhadi,
  Roesner, and Choi}]{realnews}
Rowan Zellers, Ari Holtzman, Hannah Rashkin, Yonatan Bisk, Ali Farhadi,
  Franziska Roesner, and Yejin Choi. 2019.
\newblock Defending against neural fake news.
\newblock In \emph{Advances in Neural Information Processing Systems}, pages
  9051--9062.

\bibitem[{Zhang et~al.(2018)Zhang, Dinan, Urbanek, Szlam, Kiela, and
  Weston}]{zhang2018personalizing}
Saizheng Zhang, Emily Dinan, Jack Urbanek, Arthur Szlam, Douwe Kiela, and Jason
  Weston. 2018.
\newblock Personalizing dialogue agents: I have a dog, do you have pets too?
\newblock In \emph{Proceedings of the 56th Annual Meeting of the Association
  for Computational Linguistics (Volume 1: Long Papers)}, pages 2204--2213.

\bibitem[{Zhang et~al.(2019)Zhang, Han, Liu, Jiang, Sun, and
  Liu}]{zhang2019ernie}
Zhengyan Zhang, Xu~Han, Zhiyuan Liu, Xin Jiang, Maosong Sun, and Qun Liu. 2019.
\newblock Ernie: Enhanced language representation with informative entities.
\newblock In \emph{Proceedings of the 57th Annual Meeting of the Association
  for Computational Linguistics}, pages 1441--1451.

\bibitem[{Zhu et~al.(2015)Zhu, Kiros, Zemel, Salakhutdinov, Urtasun, Torralba,
  and Fidler}]{BooksCorpus}
Yukun Zhu, Ryan Kiros, Richard~S. Zemel, Ruslan Salakhutdinov, Raquel Urtasun,
  Antonio Torralba, and Sanja Fidler. 2015.
\newblock \href {http://arxiv.org/abs/1506.06724} {Aligning books and movies:
  Towards story-like visual explanations by watching movies and reading books}.
\newblock \emph{CoRR}, abs/1506.06724.

\end{thebibliography}
\bibliographystyle{acl_natbib}

\appendix

\clearpage
\section{Appendices}
\label{sec:appendix}

\subsection{\GPTTWO~ Hyperparameters:}
\label{subsec:gpt2-hyperparams}
 We used the BPE subword tokenizer from \citet{radford2019language} to tokenize each sentence of the {\ROC} story dataset. The maximum sequence length is set to 1024 learned positional embeddings. An Adam optimizer \citep{kingma2014adam} with learning rate of 0.0001 is utilized. We added dropout to both the embedding layer and multi-head attention layers with 0.1 probability. Gradients were clipped to a global gradient norm of 5.0. 
We finetuned the {\GPTTWO} models for 5 epochs and selected the best one with the lowest perplexity on the validation set.  

\subsection{\BERT~ Hyperparameters:}
\label{subsec:bert-hyperparams}
We set the maximum sequence length to 512 learned positional embeddings. We used a WordPiece tokenizer with the \texttt{bert-large-uncased} vocabulary for tokenization. The model was also optimized with an Adam optimizer with a learning rate of 0.0001, but it used a weight decay of 0.01. Gradients are clipped to a global norm of 1.0. We also added dropout to embedding layer and multi-head attention layers with 0.1 probability. For the selection of margin, we tried 0.1, 0.5, 1.0, and 5.0. The choice of 5.0 gives the best result.


\subsection{Model Size}
\label{subsec:ablation_model_size}
In addition to analyzing the effect of scale on our conditional generative model, we also performed an ablation study on the model size of our {\GPTTWO}-based keyword \textcolor{black}{predictor} and {\BERT}-based ranker models. The results in Table \ref{table:ablation_model_size} show that increasing the keyword model size from 774M to 2B reduces the repetition while keeping the other scores similar. 
Increasing the size of our contextual ranker from 336M to 1.2B reduces the repetition while also decreasing the diversity. It is not clear which one is better. 
We conjecture that as the positive samples, $R^i$, we used to train our contextual ranker are weakly supervised, and the fact that we used templates to synthetically convert knowledge triples to knowledge sentences, scaling the model size might be overfitting to noise.
We, therefore, use the smaller, more computationally efficient model with 336M parameters for ranker models in all our experiments.

\begin{table}[!htb]
\centering\scalebox{0.8}{
\begin{tabular}{l@{\hskip3pt}cccccc}
\hline 
Name ({\it a-b-c}) & PPL $\downarrow$  & Repeat $\downarrow$ & Distinct $\uparrow$ \\
\hline\hline
2B-2B-336M & {\textbf{6.31}} 	& 21.2	& {\textbf{82.6}}  \\
2B-2B-1.2B &	6.35	& {\textbf{19.7}}	& 81.2 \\
2B-774M-1.2B & 6.33 	& 20.4	& 81.4 \\
\hline
\end{tabular}
}
\caption{\label{table:ablation_model_size} Ablation studies varying keyword prediction model ({\it b}) and ranking model ({\it c}) keeping the conditional generator fixed ({\it a}). Increasing keyword prediction model reduces repetition. Larger ranking models does not give consistently better scores as it may overfit to noise due to the weakly supervised labels. }
\vspace{-3mm}
\end{table} 

\subsection{Datasets Used for pre-trained Models}
\label{subsec:pre-trained-models}
The pre-trained {\GPTTWO} models were trained on a 174GB corpora including: Wikipedia \cite{bert}, OpenWebText \cite{openwebtext}, RealNews \cite{realnews}, and CC-Stories \cite{ccstories}.
For BERT models we include BooksCorpus \cite{BooksCorpus} in the training dataset.

\subsection{Input Format}
\label{subsec:input-representations}
For the format of $R^j$, we add a `` SEP " string to separate different knowledge sentences $r^j_{k}$ in $R^j$. We add a `` EOK " string to denote the end of knowledge sentences.

For the story context $X^{i-1} = \{x^1, x^2, \cdots, x^{i-1}\}$ where $x^j = [R^j, s^j]$, we add a ``OS" token to the end of each $s^j$ to denote the end of sentence. At the end of the story, a  `` $|$endoftext$|$ " token is appended.  We then concatenated $X^{i-1}$ with $R^i$ as the input to conditional generator.

\subsection{Preprocessing}
\label{subsec:preprocessing}
We follow the implementation of \citet{guan2020knowledge}, to replace `` ." with ``. ". More details have already been provided in Section \ref{subsec:datasets}.

\subsubsection{Computation Resources}
\label{subsec:compute-resource}
We used up to 20 DGX-1V servers (a total of 160 Tesla V100 32GB GPUs) for our experiments.
Our infrastructure is highly optimized for large scale deep learning applications. The servers are connected via NVSwitch with 300 GB/sec bandwidth between GPUs inside a server and 100 GB/sec of interconnect bandwidth between servers using 8 InfiniBand adapters per server. As our story dataset is small, the training of {\GPTTWO} models usually takes less than three hours. For the {\BERT} model, it took eight hours as we have $\sim$ 15 million samples to train.

\subsection{Generation Examples}
\label{subsec:generation-examples}
\subsubsection{Single Sentence Context}
We present five generation examples keeping the context  fixed to one sentence.

\begin{table}[!htb]
\centering\scalebox{0.9}{
\begin{tabular}{|@{\hskip3pt}p{8cm}|}
\hline
  \color{blue}{Context:} { [MALE] and his friends are bored on a weekend day.}
    \\ \hline
 \color{blue}{\citet{guan2019story}} \\
  {they decided to sneak into a movie . they got a bunch of popcorn and a movie . then , they made it to a movie . now , they have a really good time .} \\ \hline  
 \color{blue}{\OurIIVIIm} {\color{red}{keyword: go, decides}}\\
 {it is going to be a big night.  [MALE] and his friends decide to go out.  they get drunk and have a great time.  when they are done they feel happy. } \\ \hline
\color{blue} {\OurVIIIb} {\color{red}keyword: go, decides} \\
 {they decided to go to the video game store. [MALE] picks out a new video game. [MALE] and his friends have fun playing the new video game. [MALE] is no longer bored.} \\ \hline
 \color{blue}{\OurVIIIb}-{\ANTONYM} {\color{red}keyword: come} \\
 {a new tv show begins to come on the local channel.  [MALE] is excited to watch his favorite tv show.  he watches his show and they all laugh together.  [MALE] and his friends have a fun time watching their favorite show.} \\ \hline  
\end{tabular}}
\caption{Example 1}
\end{table}


\begin{table}[!htb]
\centering\scalebox{0.9}{
\begin{tabular}{|@{\hskip3pt}p{8cm}|}
\hline
  \color{blue}{Context:} {[MALE] had a heavy duty backpack. }
    \\ \hline
 \color{blue}{\citet{guan2019story}} \\
  {he took it to the airport . he checked his pockets . [MALE] did not have a bag . [MALE] was sad .} \\ \hline  
 \color{blue}{\OurIIVIIm} {\color{red}{keyword: took}}\\
 {he took it to the store one morning.  he put it on the shelf at the store.  a coworker looked at it.  then he told [MALE] he hated lugging around. } \\ \hline
\color{blue} {\OurVIIIb} {\color{red}keyword: took} \\
 {he took it to the army base camp. he packed it with items like food , ammo , and tools. he took it to the jungle every day. one day , he took it out and his backpack exploded.} \\ \hline
 \color{blue}{\OurVIIIb}-{\ANTONYM} {\color{red}keyword: abstain} \\
 {he abstained from buying another backpack.  [MALE] used the old backpack for years.  his old backpack was a big burden.  [MALE] bought a new backpack with greater weight capacity.} \\ \hline  
\end{tabular}}
\caption{Example 2}
\end{table}

\begin{table}[!htb]
\centering\scalebox{0.9}{
\begin{tabular}{|@{\hskip3pt}p{8cm}|}
\hline
  \color{blue}{Context:} {[NEUTRAL] 's blind date insisted on going to the new sushi place. }
    \\ \hline
 \color{blue}{\citet{guan2019story}} \\
  {she accepted without saying a word . later , he picked out a sushi roll . [NEUTRAL] was so happy with her new sushi rolls . she had to wait for him to come back .} \\ \hline  
 \color{blue}{\OurIIVIIm} {\color{red}{keyword: sure, date}}\\
 {she decided to go to the restaurant.  she ordered a blowfish.  they gave her a small grin.  she had a great time.  } \\ \hline
\color{blue} {\OurVIIIb} {\color{red}keyword: thought, rude} \\
 {[NEUTRAL] was nervous because he was worried she would be rude. but at the table , she was sweet and friendly. and she was very nice , even though she ate little sushi. [NEUTRAL] wondered if maybe blind dates were actually very wonderful !} \\ \hline
 \color{blue}{\OurVIIIb}-{\ANTONYM} {\color{red}keyword: uncertain} \\
 {[NEUTRAL] was uncertain.  she was not a sushi person but agreed to try it.  [NEUTRAL] and her blind date ordered salmon sushi.  the salmon sushi was delicious. } \\ \hline  
\end{tabular}}
\caption{Example 3}
\end{table}


\begin{table}[!htb]
\centering\scalebox{0.9}{
\begin{tabular}{|@{\hskip3pt}p{8cm}|}
\hline
  \color{blue}{Context:} { [MALE] hiked up the local hill. }
    \\ \hline
 \color{blue}{\citet{guan2019story}} \\
  {he stopped and picked up his brush . he used that brush to brush his teeth . afterwards , he was ready to ride back home . he went on his way home .} \\ \hline  
 \color{blue}{\OurIIVIIm} {\color{red}{keyword: top}}\\
 {he got a great view from the top.  his legs felt great.  he climbed up the mountain and felt much better.  he went back to the hill and reached the top. } \\ \hline
\color{blue} {\OurVIIIb} {\color{red}keyword: top} \\
 {he was at the top of the hill looking down. the weather looked beautiful. the clouds suddenly passed. the sun began to shine over the land again.} \\ \hline
 \color{blue}{\OurVIIIb}-{\ANTONYM} {\color{red}keyword: bottom} \\
 {he saw a huge boulder at the bottom.  he could n't believe its size.  he had to climb to the top of the hill.  he was afraid he would fall if he stood on top. } \\ \hline  
\end{tabular}}
\caption{\label{tab: example_4} Example 4}
\end{table}

\begin{table}[!htb]
\centering\scalebox{0.9}{
\begin{tabular}{|@{\hskip3pt}p{8cm}|}
\hline
  \color{blue}{Context:} {her favorite glasses were ruined.}
    \\ \hline
 \color{blue}{\citet{guan2019story}} \\
  {she was really upset . she had no choice but to buy new ones . she went to the mall to shop and found a great pair . she bought them and was so happy .} \\ \hline  
 \color{blue}{\OurIIVIIm} {\color{red}{keyword: wore}}\\
 {she wore them every day.  then she had a horrible accident.  her sister had broken them.  she had to buy new ones. } \\ \hline
\color{blue} {\OurVIIIb} {\color{red}keyword: wore} \\
 {she wore them daily. she got them at the shoe store. she found them and they were her favorite. after a while they got broken.} \\ \hline
 \color{blue}{\OurVIIIb}-{\ANTONYM} {\color{red}keyword: refresh} \\
 {she was desperate to refresh them.  she found a recipe online.  she poured bleach in them.  she was glad to have functional glasses. } \\ \hline  
\end{tabular}}
\caption{Example 5}
\end{table}

\vspace*{40px}

\subsubsection{Varying Number of Sentences in the Context}
We present three generation examples varying the number of sentences in the context.

\begin{table}[h]
\centering\scalebox{0.9}{
\begin{tabular}{|@{\hskip3pt}p{8cm}|}
\hline
  \color{blue}{Context:} {[MALE] and his friends are bored on a weekend day.  it is going to be a big night. }
    \\ \hline
 \color{blue}{\OurIIVIIm} {\color{red}{keyword: go, decides}}\\
 { [MALE] and his friends decide to go out.  they get drunk and have a great time.  when they are done they feel happy. } \\ \hline
\color{blue} {\OurVIIIb} {\color{red}keyword: decides} \\
 {[MALE] decides to watch horror movies. his friends join him. they watch horror movies all night.} \\ \hline
 \color{blue}{\OurVIIIb}-{\ANTONYM} {\color{red}keyword: stop} \\
 {they stop by a local arcade after the baseball game. [MALE] turns on his favorite arcade game. [MALE] 's favorite game makes him feel happy.} \\ \hline  
\end{tabular}}
\caption{\label{tab: example_14} The context contains {\color{blue}{two}} sentences. }
\end{table}

\begin{table}[h]
\centering\scalebox{0.9}{
\begin{tabular}{|@{\hskip3pt}p{8cm}|}
\hline
  \color{blue}{Context:} { [MALE] liked trading commodities.  one day [MALE] was trading several products.  [MALE] noticed that he gained a lot of money from the new trades. }
    \\ \hline
 \color{blue}{\OurIIVIIm} {\color{red}{keyword: lost, money}}\\
 { [MALE] invested the money into stocks.  [MALE] lost all of his trading profits.} \\ \hline
\color{blue} {\OurVIIIb} {\color{red}keyword: realize} \\
 {[MALE] realized that the new trades were his best trading idea. [MALE] quit his old job and went full time on his new trading idea.} \\ \hline
 \color{blue}{\OurVIIIb}-{\ANTONYM} {\color{red}keyword: keep} \\
 {[MALE] decided to keep doing this.  now [MALE] had a successful trading career.} \\ \hline  
\end{tabular}}
\caption{\label{tab: example_15} The context contains {\color{blue}{three}} sentences. }
\end{table}

\begin{table}[h]
\centering\scalebox{0.9}{
\begin{tabular}{|@{\hskip3pt}p{8cm}|}
\hline
  \color{blue}{Context:} {we went to a dinner and show one night. the first guy we saw was really nice. after that we asked him out. he agreed.}
    \\ \hline
 \color{blue}{\OurIIVIIm} {\color{red}{keyword: got}}\\
 { we got into a bad argument.} \\ \hline
\color{blue} {\OurVIIIb} {\color{red}keyword: NONE (no keywords)} \\
 {we had a really good time.} \\ \hline
 \color{blue}{\OurVIIIb}-{\ANTONYM} {\color{red}keyword: end} \\
 {we ended up dating exclusively.} \\ \hline  
\end{tabular}}
\caption{The context contains {\color{blue}{four}} sentences. }
\end{table}

\vspace*{100px}
\subsection{Human Evaluation Interface for Annotators}
\label{subsec:amt-interface}
Below we provide the interfaces used for human evaluations (coherence, fluency, logical consistency, and controllability).

\begin{figure*}[!htb]
\begin{center}
    \scalebox{0.8}{
        \includegraphics[width=\linewidth]{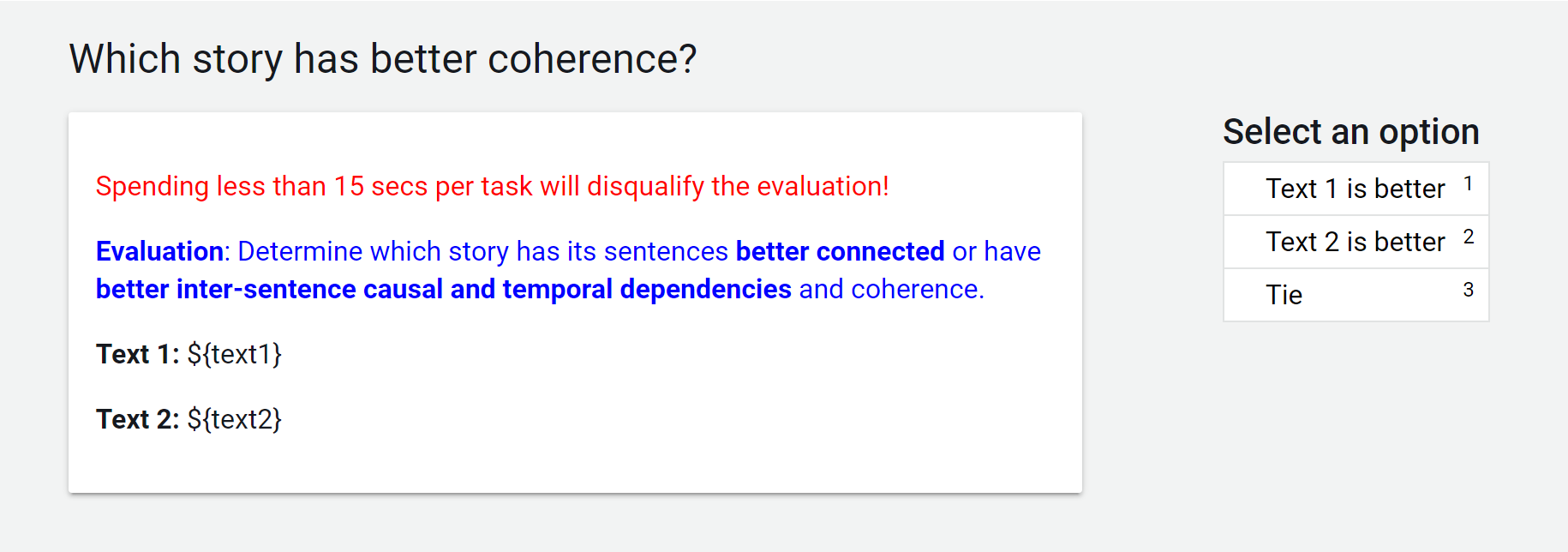}
    }
\end{center}
\end{figure*}

\begin{figure*}[!htb]
\begin{center}
    \scalebox{0.8}{
        \includegraphics[width=\linewidth]{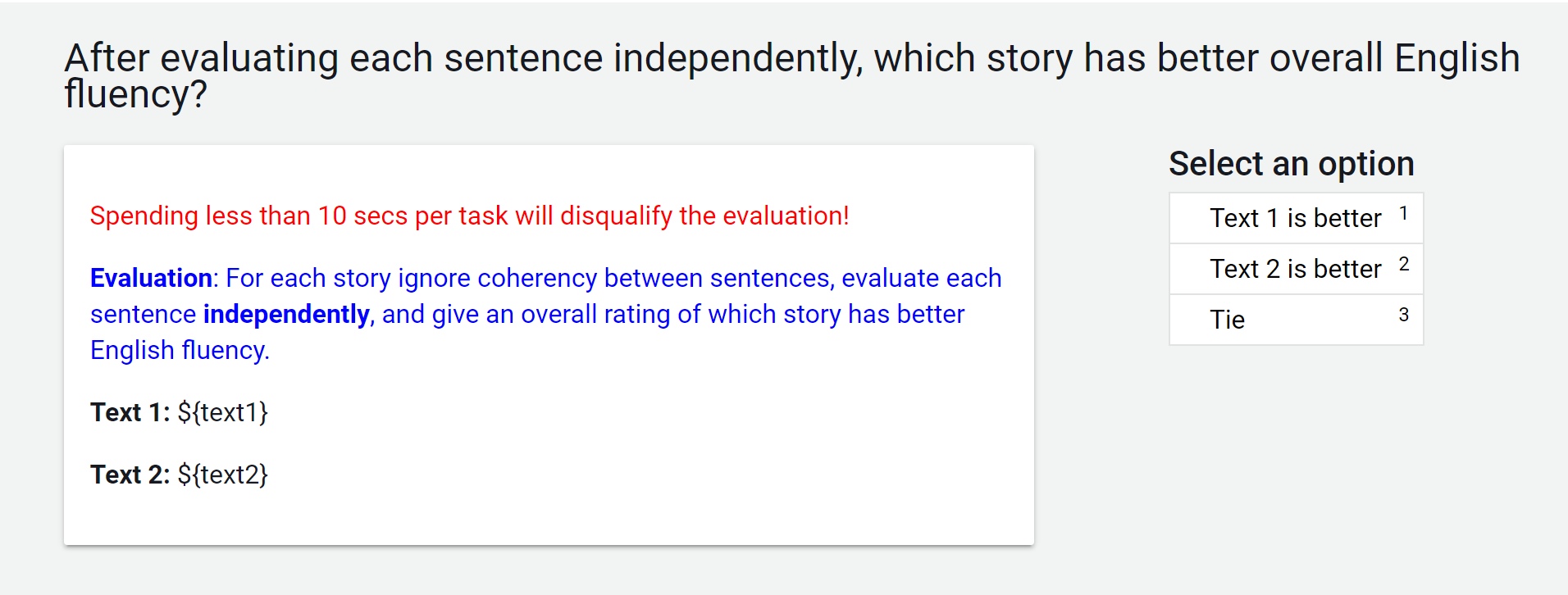}
    }
\end{center}
\end{figure*}

\begin{figure*}[!htb]
\begin{center}
    \scalebox{0.8}{
        \includegraphics[width=\linewidth]{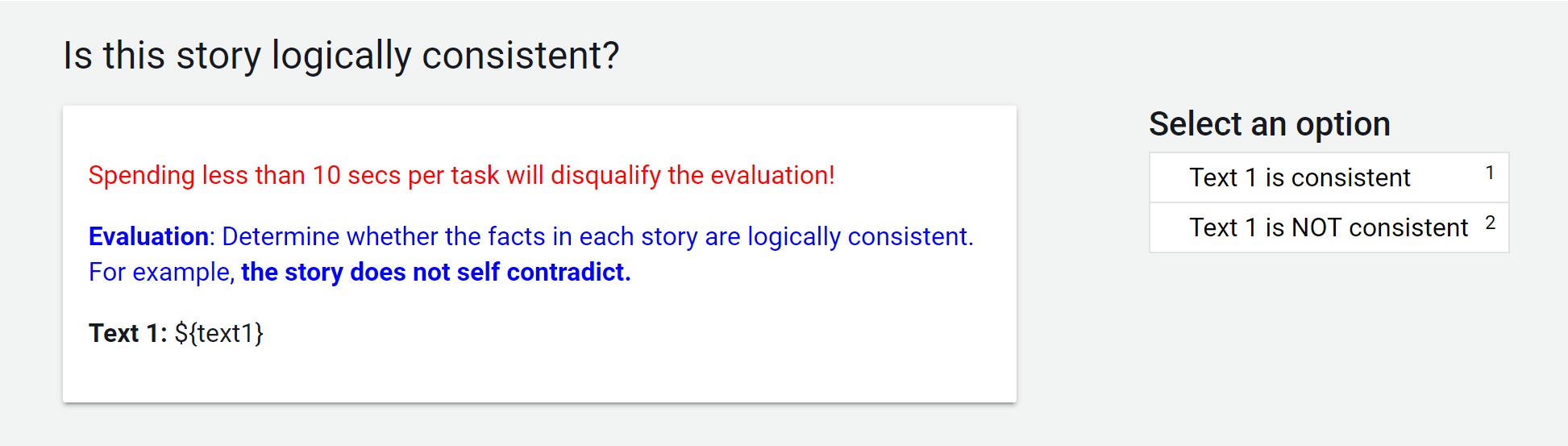}
    }
\end{center}
\end{figure*}

\begin{figure*}[!htb]
\begin{center}
    \scalebox{0.8}{
        \includegraphics[width=\linewidth]{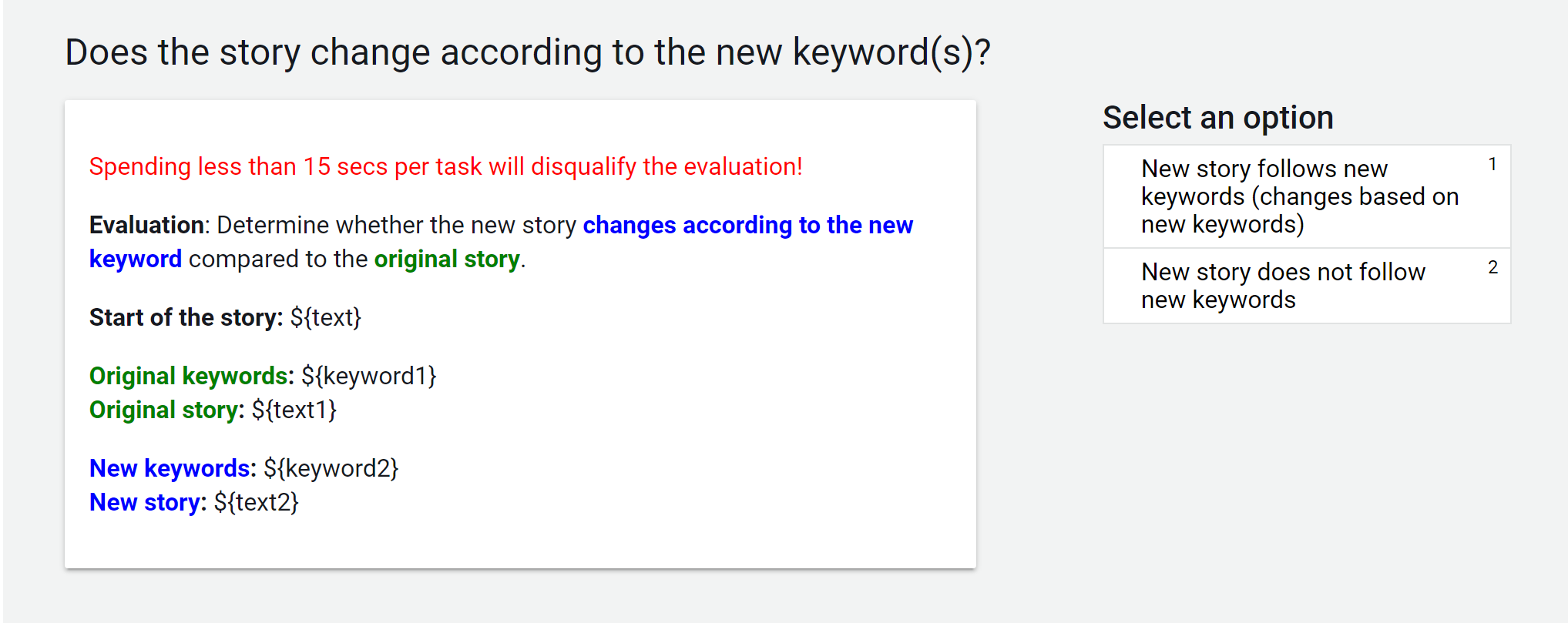}
    }
\end{center}
\end{figure*}


\end{document}